\documentclass[runningheads]{llncs}
\usepackage{graphicx}
\usepackage{bm}
\usepackage{amsfonts}
\usepackage{multirow}
\usepackage[linesnumbered, ruled, vlined]{algorithm2e}
\usepackage{amsmath}
\usepackage{algorithmic}
\usepackage{xcolor}

\newcommand\blfootnote[1]{%
  \begingroup
  \renewcommand\thefootnote{}\footnote{#1}%
  \addtocounter{footnote}{-1}%
  \endgroup
}
\begin{document}
\title{Combining Safe Interval Path Planning and Constrained Path Following Control: Preliminary Results \thanks{This work was partially supported by RFBR Grant no. 18-37-20032 and by the ``RUDN University Program 5-100''}}
\titlerunning{Combining Safe Interval Path Planning and Constrained Path Following ...}
%
\author{Konstantin Yakovlev\inst{1, 2, 5}\orcidID{0000-0002-4377-321X} \and
Anton Andreychuk\inst{1,3}\orcidID{0000-0001-5320-4603} \and
Julia Belinskaya\inst{1,4}\orcidID{0000-0002-4136-7912} \and
Dmitry Makarov\inst{1,5}\orcidID{0000-0001-8930-1288}
}
\authorrunning{K. Yakovlev et al.}
%
\institute{Artificial Intelligence Research Institute, Federal Research Center "Computer Science and Control" of Russian Academy of Sciences, Moscow, Russia
\url{http://www.rairi.ru/}
\and
National Research University Higher School of Economics, Moscow, Russia
\and
Peoples' Friendship University of Russia (RUDN University), Moscow, Russia
\and
Bauman Moscow State Technical University, Moscow, Russia
\and
Moscow Institute of Physics and Technology, Dolgoprudny, Russia\\
\email{yakovlev@isa.ru, andreychuk@mail.com, makarov@isa.ru, belinskaya.us@gmail.com}
}
\maketitle              
\begin{abstract}
We study the navigation problem for a robot moving amidst static and dynamic obstacles and rely on a hierarchical approach to solve it. First, the reference trajectory is planned by the safe interval path planning algorithm that is capable of handling any-angle translations and rotations. Second, the path following problem is treated as the constrained control problem and the original flatness-based approach is proposed to generate control. We suggest a few enhancements for the path planning algorithm aimed at finding trajectories that are more likely to be followed by a robot without collisions. Results of the conducted experimental evaluation show that the number of successfully solved navigation instances significantly increases when using the suggested techniques. \blfootnote{\textit{Camera-ready version of the paper as submitted to ICR'19}}

\end{abstract}
\section{Introduction}
Moving from one location to the other without collisions is one of the fundamental problems in mobile robotics. Two approaches to solve it are common: reactive and deliberative. Methods following reactive approach, e.g. BUG algorithms \cite{ng2007performance}, \cite{magid2004cautiousbug} or ORCA \cite{van2011reciprocal}, rely on minimum knowledge and on simple ``follow-straight-line-to-the-goal'' strategy combined with a fixed set of rules to avoid collisions. Deliberative methods utilize knowledge about the environment to plan the collision-free trajectory avoiding detours and/or deadlocks. In this paper we follow the second approach.

When environment is static, it is common to solve a discretized version of the problem, e.g. treat the path finding as a graph search problem. This graph can incorporate information about the kinematic/dynamic constraints of the robot. In this case utilizing one of the family of the RRT planners \cite{lavalle2001randomized} is widespread. The other option is to plan for a path in a simplified graph which models only the environment, e.g. occupancy grid \cite{yap2002grid} or visibility graph \cite{wooden2006graph}, and then use this path as a reference trajectory the robot has to follow \cite{nieuwenhuisen2016local}. In this work we adopt the latter approach.

Presence of dynamic obstacles adds another layer of complexity to path finding as one needs to reason about the time as well. Typically timeline is discretize into the timesteps and robot's moves are restricted to last exactly one timestep. In this setting, conventional heuristic search, e.g. A* \cite{hart1968formal} or one of its descendants, might be run to solve the problem. To make the search more efficient it is reasonable to group the time steps into the intervals as suggested by the paradigm of Safe Interval Path Planning (SIPP) \cite{phillips2011}.

In our work we do not restrict all robot's moves to last the same and utilize a modification of SIPP, i.e. AA-SIPP \cite{yakovlev2017aasipp}, that allows following not only edges that were initially present in the graph but also the newly build ones that represent the shortcuts. Original AA-SIPP as described in \cite{yakovlev2017aasipp} is supposed to handle only translation moves, while we wish to handle turn-in-place moves as well. Thus we propose an appropriate extension of the algorithm.

To follow the planned collision-free trajectory flatness-based approach is used. Many models of real vehicles can be defined as a differentially flat \cite{fliess1995} system (see, for example, \cite{sira-ramirez2004}, \cite{sahoo2018}), i.e. the systems which are equivalent to the Brunovsky normal form \cite{isidori1995}. At this stage model constraints, e.g. maximum acceleration, are taken into account and desired admissible trajectory and control are defined (see \cite{chetverikov2004},\cite{chetverikov2007}, \cite{belinskaya2012} for details). Then a state feedback control providing asymptotic stabilization around the obtained admissible trajectory is constructed. It is likely that the real trajectory of closed-loop system does not exactly match the planned one, thus collisions might occur. To mitigate this issue we suggest to modify path planning algorithm in such way that it keeps additional safety margin in the time-space while planning for a trajectory. We evaluate the suggested approach empirically and show that it can definitely contribute to finding trajectories that have higher chances to be followed without collisions.

\section{Problem Statement}

\begin{figure}[t]
    \centering
    \includegraphics[width=0.7\textwidth]{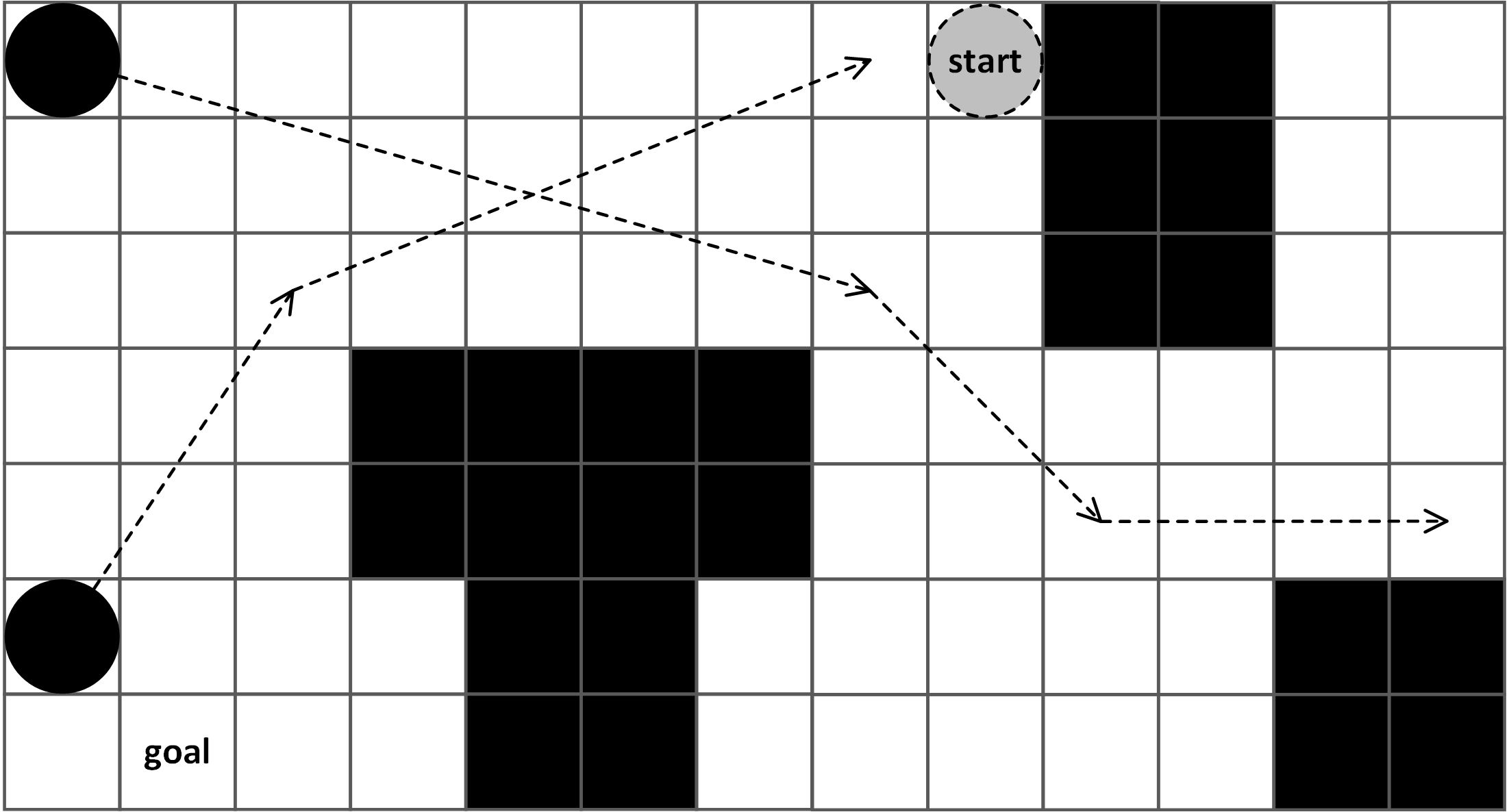}
    \caption{An example of the problem instance. Robot and dynamic obstacles are modelled as open disks. Environment is tessellated into the grid. Black cells represent static obstacles, black circles - dynamic obstacles with known trajectories marked by dashed lines. Grey circle is a robot and its task is to find a trajectory from start location to the goal one avoiding static and dynamic obstacles.}
    \label{problem-statement}
\end{figure}

We study the problem of navigating the wheeled robot in 2D in the presence of static and dynamic obstacles. Robot's workspace is a bounded rectangle comprised of the free space and the obstacles: $W=W_{free} \cup W_{obst}, W_{obst}=W \backslash W_{free}$. Workspace is tessellated to the regular square grid, composed of $W \times H$ cells. The cell of the grid is blocked if its interior contains at least one point from $W_{obst}$, otherwise the cell is un-blocked. Robot is modelled as an open disk of radius $r=0.5l$, where $l$ is the size of the grid cell. Valid locations for the robot are the centers of the un-blocked cells. Robot's state is a tuple $[x, y, \theta]$, where $x, y$ are the coordinates in meters and $\theta \in [0, 2\pi]$ in rads. The union of all valid stated is denoted as $C_{free}$.

Robot's action space is: wait in place, rotate in place, translate from one un-blocked cell to the other. Trajectory of a robot is a sequence of such actions. Formally the trajectory is a mapping 
\begin{equation}\label{def-traj}
\bm{tr}_0: [0, \infty) \rightarrow C_{free}.
\end{equation}

We suppose that each action composing the trajectory ends with a full stop. Thus one can state that: 
\begin{equation}\label{zero_velocity}
\dot{\bm{tr}_0}(t_0) = 0, \quad \dot{\bm{tr}_0}(t_f) = 0,
\end{equation}
where $t_0$ and $t_f$ are the start and finish times of an action.

Besides a robot $n$ dynamic obstacles are navigating the same configuration space. Thus, $n$ mappings defining their trajectories are also given, $\bm{tr}_i: [0, \infty) \rightarrow C_{free}$, $i=1, \ldots , n$. We assume that the obstacles do not appear/disappear, i.e. at time $t=0$ they hold their initial positions and after finishing their moves they stay in $C_{free}$. Without loss of generality we assume that the obstacles are translating-and-rotating open disks of radii $r$ and move in the same way as the robot, i.e. initially each dynamic obstacle is at some free grid cell and when translating it finishes its move at the center of another grid cell as well.

From the path planning perspective the problem is formulated as follows. Given a start and a goal positions, $s, g \in C_{free}$, find a trajectory, $\bm{tr}_0: [0, \infty) \rightarrow C_{free}$, such that \textit{i}) it is collision free w.r.t to static and dynamic obstacles, i.e. at each moment of time robot is at least $r$ units away from the closest static obstacle(s) and at least $2r$ units away from the closest dynamic obstacle(s); \textit{ii}) it starts and finishes appropriately, i.e. $\bm{tr}_0(0) = s$, and $\exists T_0: \bm{tr}_0(t)=g, \forall t \geq T_0$. For planning purposes we assume that robot accelerates/decelerates instantaneously, i.e. each segment of $\bm{tr}_0$ defines a uniform linear or angular motion.


After $\bm{tr}_0$ is constructed by a planner we need to solve a path following problem, i.e. construct a control that will follow the prescribed trajectory. To do so we suppose that a robot model is differentially flat, i.e. it is equivalent to the Brunovsky normal form, given as follows
\begin{equation}\label{sys_model}
\ddot{\bm{x}} = \bm{u}(\bm{x},t), \quad \bm{x}(0)=\bm{x}^0,
\end{equation}
where $\ddot{\bm{x}} = d^2\bm{x} / dt^2$ denotes the second time derivative, $\bm{x} = [x,y,\theta] \in \mathbb{R}^3$ is a state vector, $x$, $y$ are the spatial coordinates, $\theta$ is the yaw angle, $\bm{x}^0$ is an initial state, $\bm{u} \in \mathbb{R}^3$ is a control that has to be determined. The model has constraints on maximum linear velocity and acceleration given as follows
\begin{equation}\label{sys_constraints}
||\bm{v}(t)||= ||\dot{\bm{x}}_{sp}(t)|| \leq v_{\mbox{max}} > 0, ||\bm{a}(t)||= ||\ddot{\bm{x}}_{sp}(t)|| \leq a_{\mbox{max}} > 0, \quad \forall t \geq 0,
\end{equation}
where $\bm{x}_{sp}(t)= [x,y] \in \mathbb{R}^2$ is a spatial coordinates vector. We use a second order of Brunovsky normal form to ensure controllability on acceleration. This type of model is rather common for mechanical systems. We also suppose that a system initial state is defined as $\bm{x}^0=\bm{tr}_0(0)$. 

Although it was assumed during the planning that vehicle accelerates and decelerates instantaneously, this is impossible due to the constraints (\ref{sys_constraints}). Therefore, the prescribed trajectory $\bm{tr}_0$ is refined taking into account these constraints. To make the refined trajectory $\bm{x}^*(t) = \bigl[x^*(t), y^*(t), \theta^*(t)\bigr]$ close to the original one, we assume that the spatial movement on each segment of $\bm{x}^*(t)$ occurs in three stages: a highest possible acceleration to required velocity, a uniform motion with constant speed and a highest possible deceleration to a full stop.

Thus, the problem of path following control is formulated as follows. We need to find an admissible control $\bm{u}(\bm{x},t)$ providing asymptotic stability of the vehicle model (\ref{sys_model}) around $\bm{x}^*(t)$ under constraints (\ref{sys_constraints}).     

\section{Method}
\subsection{Path planning}
We plan collision-free trajectories with the any-angle safe interval path planner enhanced to handle rotate-in-place actions. We dub planner AAt-SIPP. 

AAt-SIPP relies on heuristic search and its search space consists of states (nodes), $s$, which are identified by tuples $s=[cfg, interval]$, where $cfg=(\textbf{pos}, \theta)$ accounts for robot's position and heading, $interval$ - is the contiguous period of time for a configuration, during which there is no collision and it is in collision one time point prior and one time point after the period. Additional data is associated with each state: $g(s), h(s), parent(s)$. $g(s)$ is the earliest possible arrival time the configuration can be reached via $parent(s)$, $h(s)$ is the consistent heuristic estimate of time needed to reach the goal-state from $s$.

On each step AAt-SIPP chooses the node with the lowest $g(s)+h(s)$ value, i.e. $f$-value, to expand. Expansion involves successors generation and updating the set of nodes constituting the fringe of the search-space -- $OPEN$. The algorithm stops either when $OPEN$ is exhausted or when the goal-state is selected for expansion. In the latter case feasible trajectory is reconstructed using backpointers $(parent(s)$.

\SetKwProg{Fn}{Function}{}{}
\SetArgSty{textnormal}
\SetInd{0.4em}{0.4em}
\SetAlFnt{\small}
\algsetup{linenosize=\tiny}
\begin{algorithm}[t]
\caption{AA-SIPP with turns}
$g(s_{start})=0$; $OPEN=\oslash$\;
insert $s_{start}$ into $OPEN$ with $f(s_{start})=h(s_{start})$\;
\While{$s_{goal}$ is not expanded}
{
    $s:=$ state with the smallest $f$-value in $OPEN$\;
    remove $s$ from $OPEN$\;
	\For{ each $cfg$ in $NEIGHBORS$($s.cfg$)}
	{
		$successors:=\text{getSuccessors}(cfg,s)$\;
		$cfg' := cfg$ reachable from $parent(s)$\;
		\If{$cfg'$ exists}
		{
			$successors=successors\cup\text{getSuccessors}(cfg',parent(s))$\;
		}
		\For{each state $s'$ in $successors$}
		{
		    add\_to\_OPEN := true\;
				\For{each visited state $s''$ such that $s''.cfg.\textbf{pos}=s'.cfg.\textbf{pos}$ and $s''.interval=s'.interval$}
				{

				    \If{$g(s')\geq g(s'')$\color{blue}{$+dur_{rot}(s',s'')$}}
				    {
				        add\_to\_OPEN := false\;
				    }
				    \ElseIf{\color{blue}{$g(s'') > g(s')+dur_{rot}(s',s'')$ and }\color{black}{$s''\in OPEN$}}
				    {
				        remove $s''$ from $OPEN$\;
				    }
				}
		    \color{black}
			\If{add\_to\_OPEN = true}
			{
			    $f(s'):=g(s')+h(s')$\;
			    insert $s'$ into $OPEN$\; 
			}
		}

	}
}
\end{algorithm}

Pseudocode of AAt-SIPP is shown in Alg.1. In case one wants to implement AA-SIPP \cite{yakovlev2017aasipp} then the additional terms in lines 14 and 16, accounting for the duration of rotate actions, should be omitted. If lines 8-10 are omitted as well one ends up with the original SIPP algorithm \cite{phillips2011}.

To efficiently handling rotate actions we modified the OPEN update routine (lines 11-20) to take the durations of rotation actions into account (corresponding code portions are highlighted in blue).

To increase the chance of not colliding with dynamic obstacles when following the constructed path we suggest to add additional safety margins to safe intervals when planning, i.e. when generating the successors. To do so we consider the robot and a dynamic obstacles to be in collision not when the distances between them is less than $2r$, but rather when is is less than $2r + \delta$, where $\delta$ is the user specified parameter. 

\subsection{Path following}

We consider a series of consecutive constrained point-to-point control problems to build a control law that ensures that the system trajectory will asymptotically converges to each path segment of $\bm{tr}_0(t)$. The point-to-point control problem is the problem of finding a control $\bm{u}$ transfer a vehicle mathematical model from a given initial state $\bm{x}_0$ to a given final state $\bm{x}_f$ during a fixed time interval $T = t_f - t_0$, where $t_0$ and $t_f$ are, accordingly, a start and a finish times of a path segment. Thus, for each segment we have  
\begin{equation}\label{ptp_conditions}
\bm{x}(t_f)= \bm{tr}_0(t_f), \quad \dot{\bm{x}}(t_0) = 0, \quad \dot{\bm{x}}(t_f) = 0.
\end{equation}
The initial state $x_0$ is defined as an actual state of a robot at time $t_0$ (it may deviate from the planned state).

Due to the flatness of (\ref{sys_model}) we consider the movement along $x$ and $y$ coordinates separately and solve each of the constrained point-to-point problems in two stages.

1. We build reference (desired) trajectories [$x^*(t)$, $y^*(t)$, $\theta^*(t)$ for $x(t)$, $y(t)$ and $\theta(t)$ accordingly so that the boundary conditions (\ref{ptp_conditions}) at each path segment and the velocity and acceleration constraints (\ref{sys_constraints}) are met. The motion law of spatial coordinates consists of three phases: acceleration to the required velocity, uniform motion with constant speed and deceleration to a full stop. The motion law for the $\theta$ coordinate is defined as an unconstrained polynomial time-dependence.

2. We build the control laws in the form of state feedback, that stabilizes the state of the system around the desired trajectories $x^*(t)$, $y^*(t)$, $\theta^*(t)$.

Let's consider each of these stages.

We suppose we have a point-to-point problem in the form of (\ref{ptp_conditions}) for the $x$ and $y$ coordinates and we have constraints (\ref{sys_constraints}). We compute components of acceleration constraint along the $x$ and $y$ coordinates
\begin{equation}
a_{\mbox{max},x} = a_{\mbox{max}}\cos\theta, \quad a_{\mbox{max},y} = a_{\mbox{max}}\sin\theta.
\end{equation}

\begin{figure}[t]
    \centering
    \includegraphics[width=0.6\textwidth]{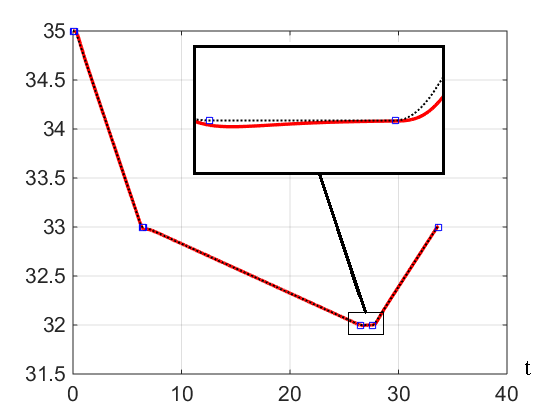}
    \caption{An example of reference trajectory $x^*(t)$ (doted curve) and real trajectory $x(t)$ (solid wide curve); squares denote boundaries of path segments.}
    \label{paths_examples}
\end{figure}

We solve each point-to-point problem similarly, therefore we consider this problem for $x$ coordinate only. Let's suppose that $x_0 \ne x_f$. If this condition
\begin{equation}\label{det_cond}
D_x = (t_f - t_0)^2 a_{\mbox{max},x}^2 - 4(x_f - x_0)a_{\mbox{max},x} \ge 0 
\end{equation}
is met, we compute the velocity of the uniform linear motion as
\begin{equation}
v_x = \left\{\begin{array}{l}
\frac{1}{2}\bigl((t_f - t_0)a_{\mbox{max},x} - \sqrt{D_x}\bigr), \quad \mbox{if }  x_f > x_0,\\
\frac{1}{2}\bigl((t_f - t_0)a_{\mbox{max},x} + \sqrt{D_x}\bigr), \quad \mbox{if } x_f < x_0.
\end{array}\right.
\end{equation}
If the condition (\ref{det_cond}) fails to satisfy, we set $v_x = v_{\mbox{max},x}$.

The motion law $x_r(t) \in \mathbb{R}$ during the acceleration we find in the space of second-order polynomials such that
\begin{equation}\label{cond_racing}
x_r(t_0) = x_0, \quad \dot{x}_r(t_0) = 0, \quad \dot{x}_r(t_0 + t_r) = v_x,
\end{equation}
where $t_r = v_x / a_{\mbox{max},x}$ is the time required for the acceleration or deceleration.
We define the uniform linear motion law as the first-order polynomial $x_s(t) = v_x(t - t_0 - t_r) + x_r(t_0 + t_r)$. The motion law $x_r(t)$ during the deceleration we find in the space of third-order polynomials such that
\begin{equation}\label{cond_deceleration}
x_d(t_f - t_r) = x_s(t_f - t_r), \quad \dot{x}_d(t_f - t_r) = v_x, \quad x_d(t_f) = x_f, \quad \dot{x}_d(t_f) = 0.
\end{equation}

Finally, we have the motion law as follows
\begin{equation}
x^{*}(t) = \left\{\begin{array}{l}
x_r(t), \quad \mbox{if } t\in [t_0,t_0 + t_r],\\
x_s(t), \quad \mbox{if } t\in(t_0 + t_r, t_f - t_r),\\
x_d(t), \quad \mbox{if } t\in [t_f - t_r,t_f]
\end{array} \right.
\end{equation}

If $x_f = x_0$, we set $x^*(t) \equiv x_0$. We chose the control law $u_x$ as the state feedback
\begin{equation}\label{control_law}
u_x(x,t) = (\lambda_1 + \lambda_2)\bigl(\dot{x}(t) - \dot{x}^*(t)\bigr) - \lambda_1\lambda_2\bigl(x(t) - x^*(t)\bigr),
\end{equation}
where $\lambda_1$, $\lambda_2$ are the roots of the characteristic equation for a differential equation for $e = x(t) - x^*(t)$. To ensure the asymptotic stability of a differential equation for $e$, the roots of the characteristic equation must be in the left half-plane of the complex plane. An example of  $x(t)$ and $x^*(t)$ trajectories is shown in the Fig. \ref{paths_examples}. 

To build the control law for the rotation we find the third-order polynomial $\theta^*(t)$ such that
\begin{equation}\label{theta_des}
\theta^*(t_0) = \theta_0, \quad \dot{\theta}^*(t_0) = 0, \quad \theta^*(t_f) = \theta_f, \quad \dot{\theta}^*(t_f) = 0.
\end{equation}
and compute the control law, similar to (\ref{control_law}). The order of polynomial is 1 less than the number of initial and final conditions.

\section{Empirical Evaluation}
Experimental evaluation was conducted in simulation on a $46 \times 70$ grid representing warehouse-like environment. The size of each cell was 1 $m^2$ and the size of the robot and the dynamic obstacles was 0.5. Translation speed was 1 m/s and rotation speed was 180 degrees per second. 128 dynamic obstacles were moving on a grid. 100 different path finding instances were generated randomly.

For the path-following algorithm it's required to set such parameters as the maximum velocity $v_{\mbox{max}}$, maximum acceleration $a_{\mbox{max}}$ and the values of roots of the characteristic equation $\lambda_1$, $\lambda_2$. The value of $v_{\mbox{max}}$ was set to 1 m/s as the same value was used for the path-planning algorithm. The values of $\lambda_1$, $\lambda_2$ were chosen empirically and were set to $-4$ and $-5$ respectively. Three different values for the acceleration rate were used: $a_{\mbox{max}}$: $5$ $m/s^2$, $8$ $m/s^2$ and $15$ $m/s^2$.

To evaluate the accuracy of the trajectory execution the root-mean-squared-error (RMSE) was computed. We computed RMSE w.r.t to the planned trajectory, as well as to the reference trajectory at the first stage of trajectory following. We have also counted the number of collisions with dynamic obstacles. The results are presented in Table 1.

Success rate in Table 1 shows the number of tasks that were completed without any collisions. As one can see in case of using the lowest acceleration speed the success rate is only 57\%. $RMSE_1$ refers to the error between the planned trajectory and the executed one. $RMSE_2$ shows the error between the reference and the executed trajectory. $RMSE_1$ and $RMSE_2$ differ significantly when $a_{\mbox{max}}$ is low, while in case when $a_{\mbox{max}}=15$ the values of $RMSE$ are much closer. This can be explained by the fact the maximum acceleration rate mostly affects the second step, when we get reference trajectories that satisfy the given boundaries of accelerations. In general one can see that executing the trajectory quite often leads to collisions. Increasing the acceleration contributes to decreasing the chance of collision, but they still occur in 27\% of cases.

To reduce the number of collisions we have used the method of inflating of collision intervals, that allows to plan trajectories with greater safety. The value of this parameter determines the additional distance that must be between the robot and the dynamic obstacle so that the algorithm considers that there is no conflict between them. There were chosen four values for the evaluation -- $0.05$, $0.1$, $0.2$ and $0.5$ meters. 

The obtained results are presented in Table 2. The values of $RMSE$ measures are not presented in this table as they all have the same trends and are almost equal to the ones presented in Table 1.

As one can see, inflating collision intervals leads to notable increase of success rate. In case of using the highest value of $a_{\mbox{max}}$ even the smallest inflating factor allows to increase the success rate up to 86\%. The best value is inflate=0.2, as higher values, i.e. 0.5, do not seem to give any positive effect. The hypothesis, why the further increasing of value of inflating parameter results to slightly worse results, is that it helps to eliminate the collisions only with dynamic obstacles, that intersect the agent's trajectory (w.r.t to the radii). As a result, the algorithm plans the trajectory in such a way, that it passes further from possibly colliding dynamic obstacles, but closer to the other ones, that were not taken into account. 

\begin{table*}[t]
\caption{Accuracy of the resultant trajectories}
\centering
{
\begin{tabular}{|c|ccc|}
\hline
Acceleration & Success Rate & $RMSE_1$ & $RMSE_2$ \\
\hline
5ms & 57\% & 0.06980 & 0.02156 \\
8ms & 69\% & 0.04705 & 0.01997 \\
15ms & 73\% & 0.03077 & 0.01878 \\
\hline
\end{tabular}
}
\end{table*}

\begin{table*}[t]
\caption{Inflating collision intervals vs success rate.}
\centering
\resizebox{\textwidth}{!}
{
\begin{tabular}{|c|cc|cc|cc|cc|}
\hline
&\multicolumn{2}{|c|}{inflate=$0.05$}&\multicolumn{2}{|c|}{inflate=$0.1$}&\multicolumn{2}{|c|}{inflate=$0.2$}&\multicolumn{2}{|c|}{inflate=$0.5$}\\
\hline
Acceleration & Success & Trajectory & Success & Trajectory & Success & Trajectory & Success & Trajectory  \\
($a_{\mbox{max}}$)& Rate & Cost & Rate & Cost & Rate & Cost & Rate & Cost\\
\hline
5ms  & 72\% & 100.3\% & 82\% & 100.96\% & 86\% & 101.45\% & 84\% & 103.7\% \\
8ms  & 82\% & 100.3\% & 86\% & 100.96\% & 87\% & 101.45\% & 85\% & 103.7\% \\
15ms & 86\% & 100.3\% & 86\% & 100.96\% & 87\% & 101.45\% & 85\% & 103.7\% \\
\hline
\end{tabular}
}
\end{table*}

The trajectory cost columns were normalized by the cost of initially planned trajectories without any inflating. As one can see in terms of trajectory cost the suggested method has a minor negative effect. There is no difference between different values of acceleration as they all performed the same trajectories, planned at the first step.

Overall, the conducted experimental evaluation has shown that suggest path-following method can produce trajectories that are very similar to the ones, obtained by the path-planning algorithm AAt-SIPP. However, in case of using low values of acceleration speed they are not collision-free in almost half of the cases in the tested scenario. To eliminate collisions we used the method of inflating of collisions intervals, that allows to increase the success rate up to 87\% while the increase of the solution cost, i.e. time spent for the trajectory following, is almost negligible.

\section{Conclusion and Future Work}
In this work we combined safe interval path planning and flatness-based constrained path following aimed at safe navigation of the differential drive robot in the environment with both static and dynamic obstacles. We have shown how to increase the chance of accomplishing the mission by a slight modification of the path planner. We have used the model flatness, polynomial approximation approach and pole placement technique to construct admissible control. Appealing direction of future research is developing of the alternatives enhancements for the path planning algorithm, e.g. taking acceleration into account, and evaluating the proposed algorithmic framework on real robots. We are also planning to apply flatness-based approach for real-life vehicle models and to develop our approach to discrete-time systems.

%
%
%
\bibliographystyle{splncs04}
\bibliography{yakovlev}
\end{document}